%% file: main.tex
\documentclass[10pt,twocolumn,letterpaper]{article}
\usepackage{cvpr}          
\usepackage[pagebackref,breaklinks,colorlinks,allcolors=cvprblue]{hyperref}
\definecolor{cvprblue}{rgb}{0.21,0.49,0.74}
\usepackage{algorithm}
\usepackage{algorithmic}
\usepackage{amsmath}
\usepackage{multirow}
\usepackage{amssymb}
\usepackage{array}
\usepackage{xurl}
\usepackage{ulem}

\usepackage{minted}
\setminted{breaklines=true}

\title{Exploiting the Final Component of Generator Architectures for AI-Generated Image Detection}

\author{Yanzhu Liu\\
Institute for Infocomm Research,\\
A*STAR, Singapore\\
{\tt\small liu\_yanzhu@a-star.edu.sg}
\and
Xiao Liu\\
Institute for Infocomm Research,\\
A*STAR, Singapore\\
{\tt\small liu\_xiao@a-star.edu.sg}
\and
Yuexuan Wang\\
Institute for Infocomm Research,\\
A*STAR, Singapore\\
{\tt\small wang\_yuexuan@a-star.edu.sg}
\and
Mondal Soumik\\
Institute for Infocomm Research,\\
A*STAR, Singapore\\
{\tt\small soumik\_mondal@a-star.edu.sg}
}

\begin{document}
\maketitle
\input{0_abstract}    
\input{1_intro}
\input{2_related}
\input{3_tax}

\input{4_method}

\input{4.1.2_detector}
\input{5_exp}
{
    \small
    \bibliographystyle{ieeenat_fullname}
    \bibliography{main}
}
\input{X_suppl}

\end{document}

%% file: 0_abstract.tex
\begin{abstract}

With the rapid proliferation of powerful image generators, accurate detection of AI-generated images has become essential for maintaining a trustworthy online environment. However, existing deepfake detectors often generalize poorly to images produced by unseen generators. Notably, despite being trained under vastly different paradigms, such as diffusion or autoregressive modeling, many modern image generators share common final architectural components that serve as the last stage for converting intermediate representations into images. Motivated by this insight, we propose to “contaminate” real images using the generator’s final component and train a detector to distinguish them from the original real images. We further introduce a taxonomy based on generators’ final components and categorize 21 widely used generators accordingly, enabling a comprehensive investigation of our method’s generalization capability. Using only 100 samples from each of three representative categories, our detector—fine-tuned on the DINOv3 backbone—achieves an average accuracy of 98.83\% across 22 testing sets from unseen generators.

\end{abstract}

%% file: 1_intro.tex
\section{Introduction}\label{sec:intro}

\begin{center}
\textit{``Last in Line, yet First to Reveal.” \\}
\end{center}
\begin{flushright}
\small{\textit{— Inspired by Sherlock Holmes}}
\end{flushright}
When examining objects like a careful detective, the last touch often reveals the most, having overwritten the layers of history beneath it. Likewise, when determining whether an image is real or AI-generated, the key is to examine the distinctive traces left by the generator, partcularly those from its final stage.
Contemporary image generators are no longer monolithic; rather, they comprise multiple modular components operating in stages. This raises a natural and important question: does the last component, which serves as the final step in producing the image, leave identifiable evidence that can be exploited for detection? Our intuition is that, regardless of the generation paradigm—generative adversarial\cite{goodfellow2014generative}, diffusion-based\cite{ho2020denoising}, autoregressive\cite{openai2024gpt4o}, or otherwise—there inevitably exists a concluding step that transforms intermediate representations into pixels, serving as the generator’s “final touch”.

We approach this question from the perspective of generation architecture. To clarify, the \textbf{architecture} refers to the structural design of a network, including its functional components and their interconnections. A network \textbf{component} is a functional module composed of multiple neural layers, such as a VAE decoder, DiT transformer, or tokenizer. In contrast, the generation \textbf{model} denotes a specific instantiation of trained weights (e.g., a pretrained network and its fine-tuned variant are considered different models). Most existing deepfake detection methods operate at the model level. They typically generate training samples from particular generation models and design sophisticated features—such as CLIP embeddings\cite{authorC2PCLIP2025}, frequency-domain cues \cite{authorAIDE2025}, multi-scale representations \cite{authorCoDE2024}, reverse-engineered traces \cite{authorDIRE2023}, or gradient-based signals \cite{authorLGrad2023}—to distinguish synthetic images from real ones. However, as new generators continue to emerge, the generalizability of these methods becomes a critical concern. When the distribution of testing data shifts significantly, these approaches often require regenerating of synthetic images from new generation models to retrain the detectors. This challenge is further amplified when users fine-tune generation models on custom data and keep the resulting models private.

Since real images are readily available from public datasets for image-related tasks (e.g., ImageNet \cite{deng2009imagenet}, MS-COCO \cite{lin2014microsoft}), we propose to “contaminate” real images with generation traces using only the generator’s final component, and subsequently train a detector to differentiate them from the original real images. Operating samples through a single component is substantially faster than executing the entire generation pipeline, particularly in diffusion models that require iterative multi-step inference. Moreover, this strategy does not necessitate the generator to be open-sourced; gray-box access to its final component suffices for implementation. 

The potential generalizability of the proposed approach is motivated by three key observations: (a) many generators share similar or even identical final components; (b) relying on only a small fraction of the total parameters reduces overfitting to specific generators; (c) the contaminated images preserve identical semantic content to their real counterparts, ensuring that the detector focuses on generation artifacts rather than semantic information. 

To comprehensively validate our hypothesis regarding the generalizability of the generator’s final component, we propose a novel taxonomy that categorizes image generators according to their final architectural component and analyze 21 widely used generators under this framework. Traditionally, image generators are grouped by generation paradigm into three major categories \cite{wei2025personalized}: Generative Adversarial Networks, Diffusion Models, and more recently, Autoregressive Models. 
However, many recent generators frequently reuse common components based on their function, such as the VAE decoder, which decodes images from latent space. 

From our taxonomy, we identify three representative final components—VAE decoder, super-resolution diffuser, and VQ de-tokenizer—and employ them to imprint traces onto real images as part of our approach. %We observe that these components correspond to distinct trace spaces. 
We conduct a series of experiments on them to investigate: whether the final component alone is sufficient to detect images generated by its corresponding full architecture; whether a detector trained on one component generalizes to images generated by architectures from other categories; and whether the detector remains effective on fine-tuned variants of the same component. Finally, we select 100 samples from each of the three components using K-medoids clustering, combining the 300 samples to fine-tune a detector built upon a pretrained DINOv3 \cite{simeoni2025dinov3} backbone with an appended fully-connected layer. The resulting detector achieves strong zero-shot performance across multiple challenging benchmarks.

The contributions of this study are as follows:
\begin{itemize}
\item We propose examining the final component of image generation architectures as a source of identifiable traces, enabling the generalizable detection.
\item We introduce a novel taxonomy of text-to-image generators based on their final architectural component, which provides insight into generalization between generators.
\item By imprinting traces onto real images and training a fine-grained, region-based detector on them, our model achieves state-of-the-art zero-shot performance across generator categories and challenging benchmarks.
\end{itemize}

%% file: 2_related.tex
\section{Related Work}
\label{sec:related}

\subsection{Architecture attribution}
For the deepfake detection task that this study focuses on, exploration from the perspective of generative architectures remains limited. While for the related task of synthesized image attribution, which aims to identify the generator responsible for a given image, few studies have investigated attribution at the architectural level.
Yang et al. \cite{yang2022deepfake} claimed to be the first to explore architecture attribution. They assessed GAN architectures and found that architectures leave globally consistent fingerprints, whereas models produce regional traces. Wi{\ss}mann et al. \cite{wissmann2024whodunit} and Xu et al. \cite{xu2024detecting} reveal that similar diffusion architectures produce similar patterns in frequency-domain features. Furthermore, Abady et al. \cite{abady2024siamese} proposed a Siamese network to verify whether two images originate from the same generation architecture, while Shahid et al. \cite{shahid2024generalized} applied one-class classification for open-set architecture scenarios.

\subsection{Last step inversion}
Few studies have examined the final step of the generation process. Laszkiewicz et al.\cite{laszkiewicz2023single} extracted features of generated images by approximating the inputs just before the final layer. The last step they focused on is a single-layer, e.g., activations, fully connected layer, or attention heads. In contrast, our study targets middle grained functional components.
\cite{wang2024trace} proposed a reverse-engineering approach to map the images back to latent features. To effective inverting the decoder, they used the output from the corresponding encoder as the initialization. While related to our sample construction method for the VAE category, their work focused on model attribution and was limited to latent-based generators.

\begin{figure*}[t]
\centering
\centering
    \includegraphics[width=1\textwidth]{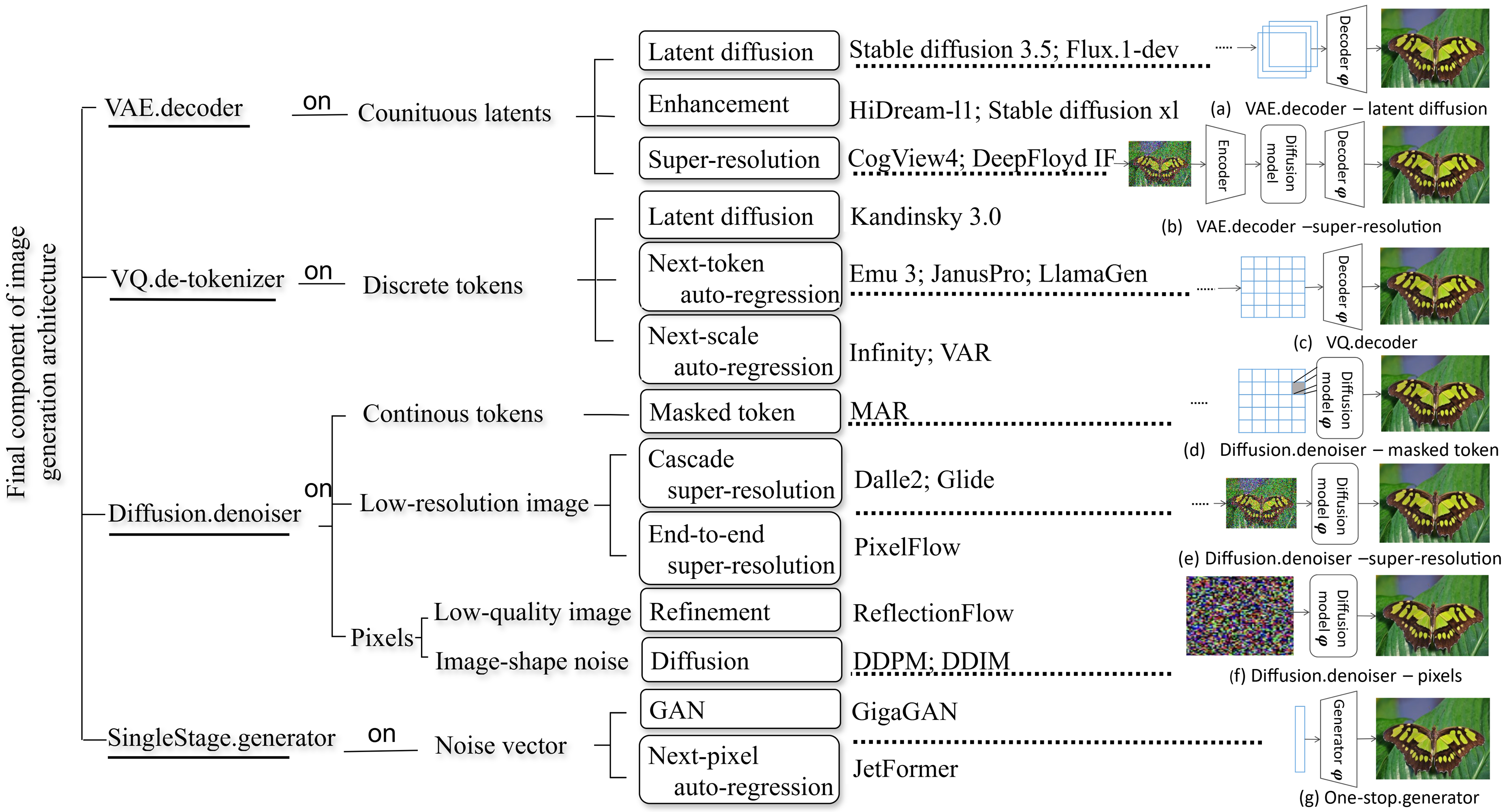}
    \caption{The proposed taxonomy of generation architectures based on final component.}
    \label{fig:2}
\end{figure*}

\subsection{Zero-shot deepfake detection}
In the zero-shot setting, the detection model is trained on samples from certain generators and applied directly to test images without fine-tuning or any prior knowledge of the generating models. Existing zero-shot methods can be broadly grouped into three categories. 
\textbf{Signal-level fingerprint construction} extracting low-level artifacts as detection cues, such as image gradients (LGrad \cite{authorLGrad2023}), pixel neighborhood differences (NPR \cite{authorNPR2024}), or generator noise patterns (DIRE \cite{authorDIRE2023}); 
\textbf{CLIP-based approaches} relying solely on CLIP image features extracted from different transformer layers (UnivFD \cite{authorUnivFD2024}, RINE \cite{authorRINE2024}), or incorporating text guidance using captions or prompts (e.g., ``real" vs. ``AI-generated") to steer classification (DeFake \cite{authorDeFake2023}, C2P-CLIP \cite{authorC2PCLIP2025}, LASTED \cite{authorLASTED2024}); 
\textbf{Robust feature design} building more resilient features, such as fusing frequency-space features (AIDE \cite{authorAIDE2025}) or learning multi-scale features through contrastive learning (CoDE \cite{authorCoDE2024}).
However, none of these methods explicitly account for the influence of the generator’s architecture or components on the generated images—a gap our approach aims to address.

%% file: 3_tax.tex
\section{A Taxonomy of Generation Architectures based on Final Component}

To facilitate the exploration of common final components in modern generation architectures and the generalizable traces they leave behind, we propose a novel taxonomy that categorizes image generators according to their final architectural component. Let $\varphi(\mathbf{z}):\mathbb{S}^{d}\rightarrow\mathbb{R}^{h\times w \times 3}$ denote this final component, where $\mathbb{S}^{d}$ represents the internal latent space. The design principles of our taxonomy, grounded in $\varphi(\mathbf{z})$, are as follows: 
\begin{itemize}
    \item \textbf{Position---Last:} The output of $\varphi(\mathbf{z})$ must be an image in RGB space $\mathbb{R}^{h\times w \times 3}$. We define the taxonomy categories based on the input $\mathbf{z}$ and its input space $\mathbb{S}^d$.
    \item \textbf{Granularity---Middle-level:} Deep network units can be considered at four levels of granularity. The finest level refers to individual layers, such as activation or pooling layers. The second level refers to blocks composed of multiple layers that serve highly detailed functions, such as upsampler in U-Net or transformer heads. The coarsest level is the entire model. ``Component" lies in the third level, representing a functional module that operates independently, such as VAE decoders or VQ de-tokenizer. 
This choice is motivated by two factors: many generators release pre-trained weights for such components separately, making them more accessible without the full pipeline; and fine-grained units, like individual layers or small blocks, leave traces that are too subtle or indistinct for reliable detection.
\item \textbf{Perspective—Implementation.} The taxonomy is based on functional implementation rather than the high-level role of the component. For example, as MAR \cite{li2024autoregressive} performs autoregressive prediction over masked tokens but ends with a diffusion denoiser, it is categorized as ``denoiser” rather than ``de-tokenizer” like other autoregressive generators. DeepFloyd IF \cite{deepfloyd2024if} performs super-resolution in latent space and ends with a VAE decoder, placing it in the decoder category rather than denoiser like other super-resolution generators.
\end{itemize} 

\begin{figure*}[t]
\centering
%\vspace{-0.2cm}
\begin{minipage}[t]{.8\textwidth}
\centering
    \includegraphics[width=.9\textwidth]{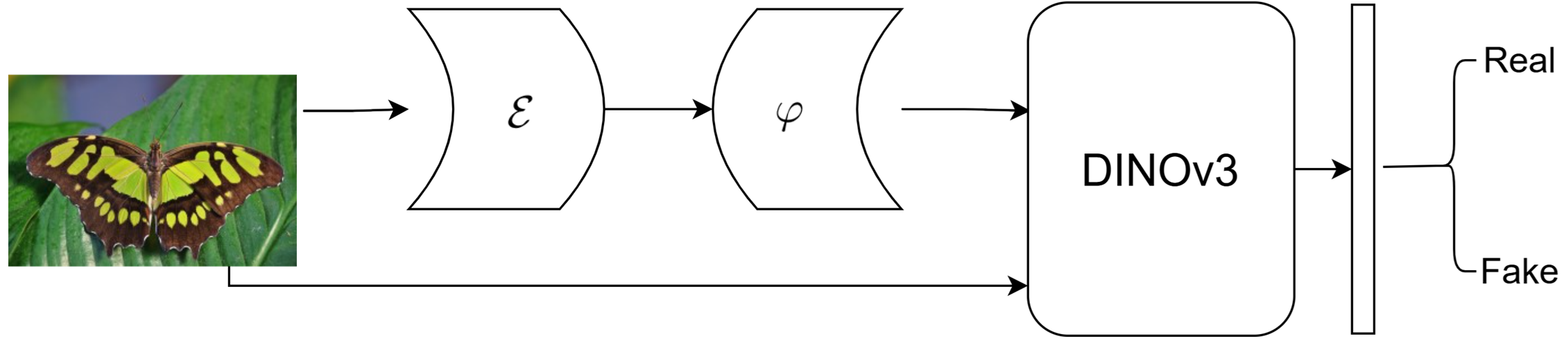}
    \caption{The proposed framework for detecting AI-generated images}
    \label{fig:1}
\end{minipage}
\end{figure*} 

As illustrated in Fig.\ref{fig:2}, we define four primary categories based on the input space $\mathbb{S}^d$, and further subdivide them according to what $\mathbf{z}$ represents, as discussed in Sections \ref{sec:3.1}-\ref{sec:3.4}.     
The categorization of all representative generators in our proposed taxonomy is based on their official source code or released documentation, with detailed references provided in the Appendix.

\subsection{VAE.decoder}\label{sec:3.1}

In the category of VAE.decoder, the final component $\phi(\mathbf{z})$ operates in a continuous latent space, i.e., $\mathbb{S}^d := \mathbb{R}^d$, where $d \ll h \times w \times 3$. It decodes a $d$-dimensional feature map or vector into the final image. As many modern generators target high-resolution outputs (e.g., $2048\times2048$), they often perform generation in a low-dimensional latent space for computational efficiency, relying on encoder–decoder structures for dimensionality reduction and expansion. 

There are three typical functional contexts in which $\mathbf{z}$ resides: (1) \textbf{Latent diffusion}, where diffusion models operate directly in the latent space, as in the Stable Diffusion family \cite{stabilityai2024sd35} and FLUX \cite{chang2024flux}. As illustrated in Fig.\ref{fig:2}(a), in this case, $\mathbf{z}$ represents the denoised latent sample predicted by the diffusion model. 
(2) \textbf{Enhancement}, where the latent representation from the diffusion model is further refined—e.g., via GAN-based distillation—to produce a more precise $\mathbf{z}$ before decoding, as in HiDream \cite{cai2025hidream}.
(3) \textbf{Latent super-resolution}, where a low-resolution image is first generated and then progressively upscaled through latent-space super-resolution stages. As shown in Fig.\ref{fig:2}(b), the low-resolution image is noised, resized to the target size, and encoded into a latent space for diffusion processing. The resulting upscaled latent, $\mathbf{z}$, is then decoded to produce the final high-resolution image. Examples include CogView 4 \cite{zheng2024cogview3} and DeepFloyd IF \cite{deepfloyd2024if}, which employ the Stable Diffusion 4×Upscaler \cite{Rombach_2022_CVPR} as the final super-resolution stage.

\subsection{VQ.de-tokenizer}\label{sec:3.2}

The final component of VQ.detokenier generators operates in discrete token space, i.e., $\mathbb{S}^d := \mathbb{Q}^p$, where $\mathbb{Q}$ is a learned codebook and $p$ denotes the number of spatial tokens (i.e., divided image patches) in the quantized  representation. 

There are three common contexts for $\mathbf{z}$: 
(1) \textbf{Next-token auto-regression}, where tokens are predicted sequentially and the VQ decoder transforms them back into image pixels, as in Emu 3 \cite{wang2024emu3}, JanusPro \cite{chen2025janus}, and LlamaGen \cite{sun2024autoregressive}. 
(2) \textbf{Next-scale autoregression}, where the generator starts from coarse-scale tokens (e.g., a $1\times1$ token map) and progressively predicts higher-resolution token maps conditioned on all previous ones, as in VAR \cite{tian2024visual} and Infinity \cite{Infinity}.
(3) \textbf{Latent diffusion hybrid}, which combines global latent diffusion for coarse image structure with autoregressive token prediction for sequentially refining discrete latent components, e.g., Kandinsky \cite{arkhipkin2023kandinsky}.  

\subsection{Diffusion.denoiser}\label{sec:3.3}

In the category of Deiffusion.denoiser, the final step is denoising, applied over three typical input spaces. (1) \textbf{Continuous token}: $\mathbb{S}^d := \mathbb{R}^p$. The image is divided into $p$ patches, with tokens represented as continuous embeddings rather than discrete codes. MAR \cite{li2024autoregressive} exemplifies this approach, as illustrated in Fig.\ref{fig:2}(d). 
(2) \textbf{Low-resolution image}: $\mathbb{S}^d := \mathbb{R}^{h_l \times w_l \times 3}$, where $h_l<h, w_l<w$. In this setting, the generator builds images progressively from low to high resolution through multiple upscaling stages, e.g., $64\times64 \rightarrow 256\times256 \rightarrow 1024\times1024$. The final step adds noise to the previous resolution’s image, resizes it to the target size, and applies diffusion denoising. This process is similar to  \textit{VAE.decoder–latent super-resolution} but operates directly in pixel space (see Fig.\ref{fig:2}(e)). Two training strategies are common: \textbf{cascade super-resolution}, where each adjacent low-to-high resolution stage is trained separately (e.g., DALL·E 3 \cite{openai2023dalle3} and GLIDE \cite{nichol2021glide}), and \textbf{end-to-end super-resolution}, where all stages are trained jointly (e.g., PixelFlow \cite{chen2025pixelflow}). 
(3) \textbf{Same-size pixels}: $\mathbb{S}^d := \mathbb{R}^{h \times w \times 3}$. The input $\mathbf{z}$ has the same spatial resolution as the target image. It can either represent a low-quality image to be refined by the diffusion denoiser, as in ReflectionFlow \cite{zhuo2025reflection}, or an initial Gaussian noise sample used in conventional diffusion models such as DDPM \cite{ho2020denoising} and DDIM \cite{songdenoising} (Fig.\ref{fig:2}(f)).            

\begin{figure*}[t]
\centering
\scriptsize
\vspace{-0.6cm}
\begin{minipage}[t]{0.3\textwidth}
\begin{algorithm}[H]
\footnotesize
\caption{Sample construction from \textit{VAE}}
\label{alg:1}
\textbf{Input:} A real image set $\mathcal{D}=\{\mathbf{x}\}$\\
\textbf{Output:} The constructed \\ image set $\{\hat{\mathbf{x}}\}$\\
\begin{algorithmic}[1]
\STATE Load $vae$ model
\FOR{ \textbf{\text{each}} $\mathbf{x}$ in $\mathcal{D}$}
\STATE $\mathbf{z}$ = vae.encode($\mathbf{x}$) 
\STATE $\hat{\mathbf{x}}$ = vae.decode($\mathbf{z}$) 
\ENDFOR
\STATE \textbf{return} $\{\hat{\mathbf{x}}\}$
\end{algorithmic}
\end{algorithm}
\end{minipage}
\hfill
\begin{minipage}[t]{0.3\textwidth}
\begin{algorithm}[H]
\small
\caption{Sample construction from \textit{VQ}}
\label{alg:2}
\textbf{Input:} A real image set $\mathcal{D}=\{\mathbf{x}\}$\\
\textbf{Output:} The constructed \\ image set $\{\hat{\mathbf{x}}\}$\\
\begin{algorithmic}[1]
\STATE Load $vq$ model
\FOR{ \textbf{\text{each}} $\mathbf{x}$ in $\mathcal{D}$}
\STATE $h$ =vq.encode($\mathbf{x}$)
\STATE $\mathbf{z}$ = vq.quantize($h$)
\STATE $\hat{\mathbf{x}}$ = vq.decode($\mathbf{z}$)
\ENDFOR
\STATE \textbf{return} $\{\hat{\mathbf{x}}\}$
\end{algorithmic}
\end{algorithm}
\end{minipage}
\hfill
\begin{minipage}[t]{0.36\textwidth}
\begin{algorithm}[H]
\small
\caption{Sample construction from \textit{Diffusion}}
\label{alg:3}
\textbf{Input:} Real images $\mathcal{D}=\{\mathbf{x}\}$\\
\textbf{Output:} Constructed set $\{\hat{\mathbf{x}}\}$\\[-1.2em]
\begin{algorithmic}[1]
\STATE Load $denoiser$ model
\FOR{ \textbf{\text{each}} $\mathbf{x}$ in $\mathcal{D}$}
\STATE $\mathbf{z}=\hat{\mathbf{x}}_T = \alpha\mathbf{x}+\beta\epsilon$, \\$\epsilon\sim N(\mathbf{0}, \mathbf{I})$\\
\FOR{$t = T-1$ to $0$}
\STATE $\tilde{\epsilon}$=denoiser($\hat{\mathbf{x}}_{t+1}$)
\STATE $\hat{\mathbf{x}}_{t}=$ step($\hat{\mathbf{x}}_{t+1}$, $\tilde{\epsilon}$))
\ENDFOR
\ENDFOR
\STATE \textbf{return} {$\{\hat{\mathbf{x}}_{0}\}$}
\end{algorithmic}
\end{algorithm}
\end{minipage}
\end{figure*}

\begin{algorithm}[t]

\small
\caption{Sparse sample selection}
\label{alg:4}
\textbf{Input:} Constructed image sets: $\mathcal{D}_1\cup\mathcal{D}_2\cup\mathcal{D}_3= 
\{\hat{\mathbf{x}}\}_{\text{VAE}}\cup$
$\{\hat{\mathbf{x}}\}_{\text{VQ}}\cup\{\hat{\mathbf{x}}\}_{\text{Diffusion}}$, the number of samples to be selected: $k$   \\
\textbf{Output:} The sparse constructed image set $\{\hat{\mathbf{x}}\}$\\[-1.2em]
\begin{algorithmic}[1]
\STATE Load pretrained \textit{dinov3} model \\% = torch.hub.load(``dinov3\_vith16plus")\\
\FOR{$i = 1$ to $3$}
\STATE features.append(\textit{dinov3}($\hat{\mathbf{x}}$)) for all $\hat{\mathbf{x}}\in \mathcal{D}_i$\\
\STATE dist = Euclidean\_distances(features) \\ %sklearn.metrics.euclidean\_distances(features)\\
\STATE $\mathcal{M}_i$= Kmedoids(dist, $k$)%sklearn.kmedoids.fasterpam(dist, $k$) 
\qquad \qquad //a set of $k$ instances of $\hat{\mathbf{x}}$
\ENDFOR
\STATE \textbf{return} $\mathcal{M}_1\cup\mathcal{M}_2\cup\mathcal{M}_3$
\end{algorithmic}
\end{algorithm}
%\end{figure}

\subsection{SingleStage.generator}\label{sec:3.4}
In the setting of SingeStage.generator, $\mathbb{S}^d := \mathcal{N}(\mathbf{0}, \mathbf{I}_d)$. The generator directly maps a sampled noise vector to the final image, as shown in Fig.\ref{fig:2}(g). This category includes GAN-based approaches such as GigaGAN \cite{kang2023gigagan}, which produce high-resolution images through large-scale adversarial training, and pixel-wise auto-regressive models like JetFormer \cite{tschannen2024jetformer}, which generate images sequentially in an end-to-end fashion.

%% file: 4_method.tex
\section{Method}\label{sec:method}

We aim to investigate whether the final component leaves identifiable traces in generated images when all preceding components are omitted, and how well such traces generalize across different generation architectures. The core idea is to use real images to elicit generation signatures and train a binary classifier on them.
Let the training set of real images be denoted as $\mathcal{D}=\{\mathbf{x}\}$, where each image $\mathbf{x}\in \mathbb{R}^{h \times w\times 3}$ is labeled as 1 (indicating a real image). Let $\varphi^*(\cdot):\mathbb{S}^d\rightarrow \mathbb{R}^{h\times w \times 3}$ represent the optimized final component of a well-trained generative model, where $\mathbb{S}^d$ is an internal space. 
To simplify notation, we do not distinguish the model’s parameter set from its corresponding mapping function, e.g., $\varphi^*$ denotes both the mapping function implemented by a network module and its learnable parameters. 

The objective function of the proposed AI-generated image detection is the standard cross-entropy on real images and reconstructed counterparts:
\begin{equation}\label{eq:1}
\mathcal{L}(\theta;\mathcal{E}) := -E_{\mathbf{x}\sim p_{\mathcal{D}}} [log(\theta(\mathbf{x})) + log(1-\theta(\varphi^*(\mathcal{E}(\mathbf{x})))]
\end{equation}
where $\mathcal{E}(\cdot):\mathbb{R}^{h\times w \times 3}\rightarrow\mathbb{S}^{d}$ maps an RGB image into the latent space. Given a test image $\mathbf{x}_t$, the classifier trained with Eq. \ref{eq:1} outputs $\theta(\mathbf{x}_t)$, indicating the probability that $\mathbf{x}_t$ is real ($1$) or AI-generated ($0$).
In this formulation, The encoder $\mathcal{E}$ determines the representation from which the corresponding fake image $\mathbf{x}$ is decoded. It is predefined and not co-trained with $\theta$. The following property formulates the intuition behind the choice of $\mathcal{E}$.

\textbf{Property 1 (Choice of encoder)} 
\textit{Let $\mathcal{E}^*$ be the encoder jointly optimized with $\varphi^*$ during generator training, i.e., $(\mathcal{E}^*,\varphi^*)=\arg\min \mathcal{L}_g(\mathcal{E}, \varphi)$, where $\mathcal{L}_g$ is the generation loss. For any encoder $\mathcal{E}$, under global optimality,
$\mathcal{L}(\theta_{\mathcal{E}^*},\mathcal{E}^*)\leq\mathcal{L}(\theta_{\mathcal{E}},\mathcal{E})$, where $\theta_{\mathcal{E}}=\arg\min_{\theta} \mathcal{L}(\theta;\mathcal{E})$.}

This property is stated under the idealized assumption of global minimizers. Nevertheless, it suggests that the encoder $\mathcal{E}^*$, being jointly optimized and distributionally aligned with the decoder $\varphi^*$, achieves the minimal attainable expected detection loss among all possible mapping functions $\mathcal{E}$.

\begin{figure*}[t]
\begin{minipage}[b]{0.42\textwidth}
\includegraphics[width=.95\linewidth]{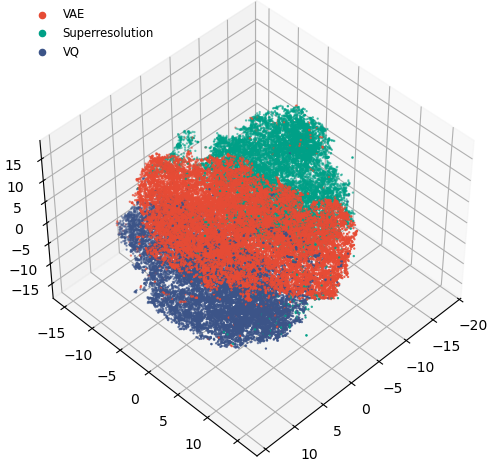}
    \caption{Feature visualization of the three final components.}
    \label{fig:6}
\end{minipage}
\hfill
\begin{minipage}[b]{0.56\textwidth}
    \includegraphics[width=.95\linewidth]{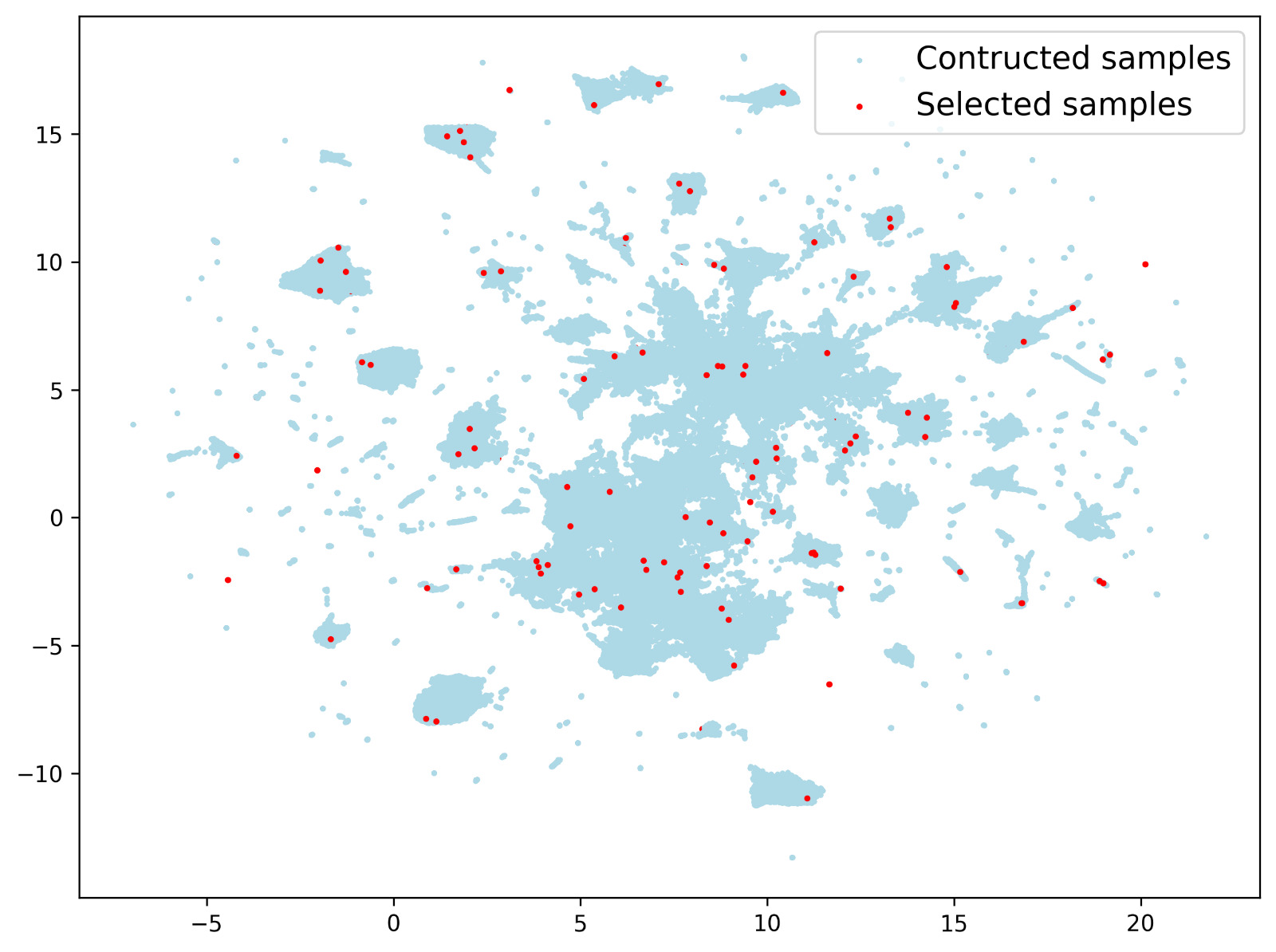}
    \caption{Sparse sample selection.}
    \label{fig:3}
\end{minipage}
\end{figure*}

\textbf{Property 2 (Non-pairwise training)} 
\textit{Let $\mathcal{D}=\{\mathbf{x}\}$ be real samples and  $\mathcal{D}'=\{\mathbf{x}'$ $= \varphi(\mathcal{E}(\mathbf{x})) \mid \mathbf{x} \in \mathcal{D}\}$ be constructed samples. Sampling mini-batches independently from $\mathcal{D}$ and $\mathcal{D}'$, rather than using paired $(\mathbf{x}_i, \mathbf{x}'_i)$, yields a gradient estimator with lower variance by SGD training, i.e., }
$\text{Var}(\hat{\nabla}_\theta^{(I)}) \le \text{Var}(\hat{\nabla}_\theta^{(P)})$,
\textit{where} \small\[
\hat{\nabla}_\theta^{(I)} = - \frac{1}{B} \sum_{\mathbf{x} \in \mathcal{B}_t} \nabla_\theta \log \theta(\mathbf{x}) - \frac{1}{B} \sum_{\mathbf{x}' \in \mathcal{B}'_t} \nabla_\theta \log \big( 1 - \theta(\mathbf{x}') \big),
\] \small\[\hat{\nabla}_\theta^{(P)} = - \frac{1}{B} \sum_{\mathbf{x} \in \mathcal{B}_t} \Big[ \nabla_\theta \log \theta(\mathbf{x}) + \nabla_\theta \log \big( 1 - \theta(\varphi(\mathcal{E}(\mathbf{x}))) \big) \Big], \]  \textit{and $\mathcal{B}_t \subseteq \mathcal{D}$,  $\mathcal{B}'_t \subseteq \mathcal{D}'$ are mini-batches of size $B$.}

A theoretical and empirical justification is provided in Appendix. This property suggests that independently shuffling real and constructed samples reduces the variance of the stochastic gradient, which in turn leads to more stable and faster convergence in SGD \cite{robbins1951stochastic,bottou2010large}. Therefore, it is advantageous to construct the negative sample set $\mathcal{D}'$ offline and sample from $\mathcal{D}$ and $\mathcal{D}'$ independently during SGD training.

Fig. \ref{fig:1} presents an overview of our method. The input is a real image, and the corresponding fake samples are constructed from it using the encoder and decoder. This operation realizes $\varphi\circ\mathcal{E}(\cdot)$ in Eq. \ref{eq:1}, leaving detectable traces while preserving the exact semantic content of the real image. The detector $\theta$ is implemented using a pretrained feature extractor (DINOv3) with an additional classification layer. In the following, we detail these two key components of the objective function in Eq. \ref{eq:1}: the fake sample construction $\varphi\circ\mathcal{E}(\cdot)$ and the detector $\theta$.

%% file: 4.1.2_detector.tex
\subsection{Fake sample construction --- $\varphi\circ$ \texorpdfstring{$\mathcal{E}$}{E}} 
\subsubsection{Construction from the three representative components}
In our taxonomy (Fig.~\ref{fig:2}), the three primary categories—VAE.decoder, VQ.de-tokenizer, and Diffusion.denoiser—are defined based on the implementation of the final component, while subcategories are further determined by the input space and functional context in which the component operates. Consequently, all components within the same primary category can share the same algorithmic pipeline for sample construction. 

For brevity, we refer to components in these categories as \textit{VAE}, \textit{VQ}, and \textit{Diffusion} respectively in the following sections. As discussed in Property 1, the encoder $\mathcal{E}^*$, which is pretrained jointly with $\varphi^*$, is used to approximate the latent input $\mathbf{z}$. Algorithm \ref{alg:1} presents the pseudocode for constructing $\varphi^*(\mathcal{E}^*(\mathbf{x}))$ using the \textit{VAE}. The implementation requires only a few lines of code and is significantly more efficient than executing the full generative pipeline.
For \textit{VQ}, the de-tokenizer acts as $\varphi^*$ and the image tokenizer as $\mathcal{E}^*$ to approximate image tokens $\mathbf{z}$, as outlined in Algorithm \ref{alg:2}.
In \textit{Diffusion}, the input $\mathbf{z}$ is a noisy version of the input, $\alpha\mathbf{x} + \beta\epsilon_t$, where $\epsilon \sim \mathcal{N}(0, I)$. The noise level can be adjusted by selecting the starting timestep $t$ for denoising (Algorithm \ref{alg:3}).

\subsubsection{Sparse sampling from combined constructions}
As the final components of the three categories differ substantially in both functionality and implementation, the constructed samples $\varphi^*(\mathcal{E}^*(\mathbf{x}))$ from each may exhibit distinct generative traces. We further investigate whether combining a small set of representative fake samples from each component can capture the overall trace space and achieve effective generalization. 
Algorithm \ref{alg:4} outlines the procedure, applying K-Medoids clustering selects $k$ representative samples from the constructed outputs of each component. The selection is performed in a reduced-dimensional feature space. Specifically, the generated samples are first encoded using a pretrained DINOv3 backbone \cite{simeoni2025dinov3}, from which 2,048-dimensional feature vectors are extracted from the final layer. K-Medoids selects $k$ samples such that most other samples are close to one of them in Euclidean distance. The selected subsets from all three components are combined to form the final training set of $\varphi^*(\mathcal{E}^*(\mathbf{x}))$ for detector training.

\subsection{Detection model — $\theta$}

Based on the Property 2, instead of loading $\mathbf{x}$ and computing $\varphi^*(\mathcal{E}^*(\mathbf{x}))$ at each training step, we directly input the real image set $\{\mathbf{x}\}$ and the constructed image set $\{\hat{\mathbf{x}}\}$ generated by Algorithms \ref{alg:1}–\ref{alg:4}, with standard shuffling applied to both sets during SGD training. Since the detector $\theta$ aims to distinguish subtle differences between real and reconstructed images, and object detection tasks inherently emphasize multi-scale spatial features, we adopt DINOv3—a foundation model pretrained for object detection—as the backbone to better capture fine-grained artifacts. A single fully connected layer is appended to the backbone, and all layers are fine-tuned during training. As our focus is on analyzing the effect of the final generative component, we employ a pretrained visual backbone with minimal modifications rather than designing a complex detector.

\begin{figure*}[t]
\begin{minipage}[b]{1\textwidth}
\centering
\addtocounter{table}{1}
%\refstepcounter{table}
\qquad \textbf{Table 1.} Detection results across different category of generators. \\[4pt]
\scriptsize
\setlength{\tabcolsep}{2pt}
\renewcommand{\arraystretch}{1.1}
\begin{tabular}{l|*{2}{l}|*{2}{l}|*{2}{l}|*{2}{l}|*{2}{l}|*{2}{l}}
\hline
\multirow{3}{*}{} & \multicolumn{6}{c|}{\textbf{diffusion.denoiser based}} & \multicolumn{4}{c|}{\textbf{vq.de-tokenizer based}} & \multicolumn{2}{c}{\textbf{vae.decoder}} \\
\cline{2-13}
& \multicolumn{2}{c|}{DALL·E 2} & \multicolumn{2}{c|}{DALL·E 3} & \multicolumn{2}{c|}{Glide} & \multicolumn{2}{c|}{Emu3} & \multicolumn{2}{c|}{LlamaGen} & \multicolumn{2}{c}{HiDream} \\
\cline{2-13}
& Acc\% & AP\% & Acc\% & AP\% & Acc\% & AP\% & Acc\% & AP\% & Acc\% & AP\% & Acc\% & AP\% \\
\hline
DIRE      & 56.45 & 67.05 & 50.35 & 53.53 & 49.50 & 49.27 & 53.65 & 56.73 & 54.05 & 59.26 & 51.40 &	51.21 \\
LGrad     & 55.65 & 55.73 & 47.65 & 40.41 & 47.45 & 51.83 & 54.80 & 48.03 & 60.70 & 63.08 & 40.05 &	39.06\\
NPR       & 49.25 & 50.18 & 50.35 & 50.48 & 48.80 & 51.43 & 51.70 & 53.05 & 52.25 & 55.48 & 47.75 &	45.51\\
FatFormer & 54.12 & 56.02 & 35.88 & 33.95 & 67.77 & 75.75 & 63.72 & 66.10 & 77.21 & 87.73 & 34.48 &	31.28\\
RINE      & 85.90 & 94.05 & 42.35 & 30.83 & 63.95 & 72.09 & 62.10 & 69.68 & 90.65 & 98.36 & 44.15 &	39.20\\
CoDE      & 54.60 & 54.26 & 73.15 & 72.76 & 80.75 & 80.36 & 77.45 & 77.06 & 90.50 & 90.11 & 71.45 &	71.06\\
C2P-CLIP  & 55.60 & 76.87 & 63.20 & 90.45 & \uwave{86.65} & \uwave{96.53} & 68.70 & 90.60 & 83.80 & 97.85 & 51.20 & 51.02\\
AIDE      & 59.75 & 42.26 & 14.15 & 31.34 & 59.75 & 76.71 & 62.50 & 56.86 & 63.25 & 84.55 & 48.30 & 37.95\\
BFree     & \uwave{93.30} & \uwave{99.16} & \uwave{96.22} & \uwave{99.64} & 68.61 & 93.81 & \uwave{99.05} & \uwave{99.97} & \uwave{99.15} & \uwave{99.98} & \uwave{81.40} & \uwave{93.19} \\\hline
Ours\_VAE & 99.60 & \textbf{100.0} & 99.70   & 99.99          & \textbf{98.85} & \textbf{99.92} & 99.70          & 99.99          & 99.70 & 99.99  & \textbf{99.15}&\textbf{99.94}         \\
Ours\_VQ  & \textbf{99.80}  & \textbf{100.0} & \textbf{99.85}          & 99.99  & 98.75        & 99.84          & 99.85          & 99.99 & 99.85 & \textbf{100.0} & 97.15	&99.51 \\
Ours\_\tiny{Diffusion}& 99.35 &99.91          & \textbf{99.85}          & \textbf{100.0} & 98.60          & 99.85          & \textbf{99.90}  & \textbf{100.0} & \textbf{99.90} & \textbf{100.0} & 88.95	&96.95\\
Ours\_Sparse& \textbf{99.80}  & 99.87          & 99.15          & 99.91          & 98.05          & 99.47          & 99.15          & 99.90  &99.15 & 99.82 &96.80	&99.21\\
\hline
\end{tabular}
\label{tab:2}
\end{minipage}
\begin{minipage}[b]{1\textwidth}
\centering 
\scriptsize
\setlength{\tabcolsep}{2pt}
\renewcommand{\arraystretch}{1.1}
\begin{tabular}{l|*{2}{l}|*{2}{l}|*{2}{l}|*{2}{l}||*{2}{l}|*{2}{l}}
\hline
\multirow{3}{*}{} & \multicolumn{12}{c}{\textbf{vae.decoder based}} \\
\cline{2-13}
& \multicolumn{2}{c|}{SD1.3} & \multicolumn{2}{c|}{SD1.4} & \multicolumn{2}{c|}{SD XL} & \multicolumn{2}{c||}{SD2} & \multicolumn{2}{c|}{Flux} & \multicolumn{2}{c}{SD 3.5} \\
\cline{2-13}
& Acc\% & AP\% & Acc\%& AP\% & Acc\% & AP\% & Acc\%& AP\%& Acc\%& AP\%& Acc\% & AP\% \\
\hline
DIRE      & 48.75 & 43.24 & 48.65 & 43.38 & 50.00 & 58.90  & 47.45 & 40.37 & 61.66 & 73.10 & 55.65 & 58.00\\
LGrad     & 54.45 & 46.46 & 55.85 & 48.14 & 58.65 & 53.26  & 48.75 & 42.80 & 81.91 & 91.81 & 77.29 & 87.08 \\
NPR       & 50.70 & 51.05 & 50.60 & 51.69 & 52.45 & 55.54  & 43.50 & 46.23 & 90.00 & 92.34 & 88.04 & 89.82 \\
FatFormer & 52.97 & 53.60 & 53.72 & 54.07 & 66.67 & 75.43  & 51.57 & 54.41 & 55.32 & 89.83 & 75.86 & 95.39 \\
RINE      & 87.60 & 95.96 & 87.15 & 95.95 & 74.60 & 85.56  & 70.30 & 80.74 & 51.70 & 86.61 & 96.49 & \uwave{99.93}\\
CoDE      & 97.75 & 97.36 & 97.30 & 96.91 & 75.20 & 74.81  & 88.05 & 87.66 & 71.09 & 72.17 & 76.35 & 75.26\\
C2P-CLIP  & 75.15 & 92.93 & 76.70 & 93.03 & 77.70 & 94.67  & 69.15 & 92.30 & 53.96 & 94.58 & 80.88 & 98.87 \\
AIDE      & 57.60 & 60.25 & 56.70 & 60.14 & 62.70 & 63.98  & 53.45 & 42.09 & \uwave{98.50} & \uwave{99.80} & 96.45 & 99.72 \\
BFree   & \uwave{99.25} & \uwave{99.99} & \uwave{99.15} & \uwave{99.99} & \uwave{99.25} & \uwave{99.98} & \uwave{99.80} & \uwave{99.93} & 69.62 & 93.62 & \uwave{98.74}  &  \uwave{99.94} \\\hline
Ours\_VAE & 99.70 & \textbf{100.0} & 99.70 & \textbf{100.0} & 99.70          & \textbf{100.0} & 99.70 & \textbf{100.0}& 99.17          & 99.98          & \textbf{99.96} & \textbf{100.0} \\
Ours\_VQ  & 99.85 & \textbf{100.0} & \textbf{99.85} & \textbf{100.0} & 99.85          & \textbf{100.0} & 99.40 &  99.90& 99.19          & 99.97          & 99.82 & 99.99        \\
Ours\_\tiny{Diffusion}& \textbf{99.90} & \textbf{100.0} & 99.80          & \textbf{100.0} & \textbf{99.90} & \textbf{100.0} & \textbf{99.85} & \textbf{100.0}& \textbf{99.89} & \textbf{100.0} & 99.86 & \textbf{100.0} \\
Ours\_Sparse& 99.15  & 99.90  & 99.15   & 99.90               & 99.15 &99.88              &  99.10 &99.80& 98.95          & 98.28          & 98.98 & 99.58 \\
\hline
\end{tabular}
\end{minipage}
\end{figure*}

%% file: 5_exp.tex
\section{Experiments}

\subsection{Model implementation \& baselines}

\textbf{Training dataset} 
The real image set used to construct the training dataset in this study is MS-COCO 2014~\cite{lin2014microsoft}, which contains 118{,}000 images depicting diverse indoor and outdoor scenes, with resolutions ranging from 384 to 880 pixels in width (or height).

To validate the generalizability of the generator’s final component, we select one representative final component from open-source generators for each of the three categories in Fig.~\ref{fig:2}: \textit{VAE.decoder}, \textit{VQ.detokenizer}, and \textit{Diffusion.denoiser}. Specifically, \textit{Stable Diffusion 2.1} represents the VAE.decoder category and is used to construct fake samples following Algorithm~\ref{alg:1}; \textit{JanusPro} represents the VQ.detokenizer category and generates samples according to Algorithm~\ref{alg:2}; and \textit{PixelFlow} represents the Diffusion.denoiser category and follows Algorithm~\ref{alg:3}. All pretrained encoders $\mathcal{E}^*$ and decoders $\varphi^*$ are detailed in the Appendix. For brevity, we refer to the corresponding training sets as \textit{VAE}, \textit{VQ}, and \textit{Diffusion} in the following discussion.

These three types of final components exhibit distinct generation traces, as visualized in Fig.~\ref{fig:6}. Features are extracted from the third layer of a pretrained DINOv3 backbone \cite{simeoni2025dinov3} and projected into three dimensions. Although the resulting distributions exhibit partial overlap, they remain visually separable. To further examine whether a small number of representative samples from each component can capture the overall trace space and achieve effective generalization, we construct an additional compact training set. Specifically, we select 100 samples from each of the \textit{VAE}, \textit{VQ}, \textit{Diffusion} sets using Algorithm \ref{alg:4} with $k=100$, and refer to it as \textit{Sparse} set in experiments. Fig.~\ref{fig:3} visualizes the selection results, where the chosen samples (highlighted in red) are distributed throughout the entire feature space.

\vspace{0.5em}
\noindent \textbf{Detector implementation}
To implement the AI-generated image detector, we adopt a pretrained DINOv3 backbone \cite{simeoni2025dinov3} %\footnote{\url{https://huggingface.co/facebook/dinov3-vith16plus-pretrain-lvd1689m}}
 with an additional fully connected layer for binary classification. All layers are fine-tuned using the standard cross-entropy loss in Eq.~\ref{eq:1} with a learning rate of $5\times10^{-7}$. %This implementation is used as the detection model in all subsequent analyses.
We train separate detectors on the training sets \textit{VAE}, \textit{VQ}, \textit{Diffusion}, and \textit{Sparse}, and evaluate them on images from different categories in our taxonomy to analyze the effectiveness of the proposed method.

\vspace{0.5em}
\noindent \textbf{Baselines}
We compare our method against nine baseline approaches: BFree~\cite{guillaro2025bias} (CVPR 2025), AIDE~\cite{authorAIDE2025} (ICLR 2025), C2P-CLIP~\cite{authorC2PCLIP2025} (AAAI 2025), CoDE~\cite{authorCoDE2024} (ECCV 2024), RINE~\cite{authorRINE2024} (ECCV 2024), FatFormer~\cite{liu2024forgery} (CVPR 2024), NPR~\cite{authorNPR2024} (CVPR 2024), LGrad~\cite{authorLGrad2023} (CVPR 2023), and DIRE~\cite{authorDIRE2023} (ICCV 2023). As these methods either did not report results on all of our test sets or used different evaluation metrics, we re-evaluate each method using its publicly released pretrained model and default hyperparameters. Notably, BFree also evaluated these baselines on a subset of our test sets; our results and conclusions regarding the baseline methods are consistent with those reported in BFree. The source code links, configurations, and pretrained models used are provided in the Appendix. Performance is assessed using two standard metrics: classification accuracy (Acc) and average precision (AP).

\begin{table*}[t]
\begin{minipage}[t]{1\textwidth}
\caption{Detection results on wild generations.}
\centering
\vspace{-0.27cm}
\scriptsize
\setlength{\tabcolsep}{2pt}
\renewcommand{\arraystretch}{1.1}
\begin{tabular}{l|*{2}{l}|*{2}{l}|*{2}{l}|*{2}{l}|*{2}{l}|*{2}{l}}
\hline
\multirow{3}{*}{} & \multicolumn{4}{c|}{\textbf{Unknown architecture}} & \multicolumn{2}{c|}{\textbf{Mixture}} & \multicolumn{6}{c}{\textbf{Wild generation}} \\
\cline{2-13}
& \multicolumn{2}{c|}{Firefly} & \multicolumn{2}{c|}{Midjourney}& \multicolumn{2}{c|}{FakeBench} & \multicolumn{2}{c|}{Reddit} & \multicolumn{2}{c|}{Facebook} & \multicolumn{2}{c}{Twitter} \\
\cline{2-13}
 & Acc\% & AP\% & Acc\% & AP\% & Acc\% & AP\% & Acc\% & AP\% & Acc\% & AP\% & Acc\% & AP\% \\
\hline
DIRE      & 50.80 & 55.06 & 53.90 & 60.56 & 44.38 & 41.72 & 57.19 & 61.93 & 56.56 & 66.02 & 47.74 & 78.72 \\
LGrad     & 47.25 & 42.41 & 57.70 & 53.09 & 44.01 & 69.44 & 57.80 & 65.31 & 66.88 & 77.26 & 40.86 & 61.63 \\
NPR       & 47.30 & 46.96 & 52.35 & 55.22 & 52.37 & 57.85 & 63.80 & 66.87 & 74.06 & 83.68 & 48.46 & 67.53 \\
FatFormer & 60.12 & 62.24 & 44.43 & 41.82 & 71.20 & 80.93 & 69.47 & 80.83 & 54.83 & 75.27 & 42.79 & 77.41 \\
RINE      & 92.10 & 99.71 & 55.45 & 59.38 & 74.77 & 97.22 & 71.20 & 81.28 & 51.56 & 55.44 & 55.93 & 62.27 \\
CoDE      & 60.65 & 60.28 & 76.90 & 76.51 & 74.75 & 73.94 & 65.80 & 61.12 & 72.19 & 67.50 & 68.41 & 78.34 \\
C2P-CLIP  & 59.45 & 82.11 & 52.85 & 69.37 & 74.77 & 91.18 & 68.40 & 78.85 & 54.37 & 71.39 & 47.27 & 75.07 \\
AIDE      & 14.40 & 31.74 & 53.45 & 40.43 & 74.78 & 86.76 & 70.87 & 82.52 & 50.00 & 55.96 & 52.78 & 59.40 \\
BFree     & \uwave{98.85} &	\uwave{99.91}  & \uwave{98.70} & \uwave{99.93}	& \uwave{80.81}	& \uwave{100.0}	 &\uwave{85.20} & \uwave{94.94 } &	\uwave{95.31 }&	9\uwave{8.98 } & \uwave{95.75 }& \uwave{99.42}  \\\hline
Ours\_VAE & \textbf{99.70} & \textbf{100.0} & \textbf{99.70} & 99.99  & \textbf{98.83} & \textbf{99.97} & 96.87  & 996.8          & 97.81          & 98.99          & 98.34          & 99.93  \\
Ours\_VQ  & \textbf{99.70} & 99.99      & 99.45     & 99.96     & 98.10       & 99.82     & 95.40      & \textbf{99.80} & 95.31    & 99.28         & 98.34          & 99.86          \\
Ours\_\tiny{Diffusion} & 98.50         & 99.96      & \textbf{99.70} & \textbf{100.0} & 94.38         & 99.78      & \textbf{98.67} & 99.64    & 97.19         & \textbf{99.46} & 97.39          & 99.98    \\
Ours\_Sparse & 98.00       & 99.73      & 99.10          & 99.90          & 97.90          &99.9      & 97.60           & 99.51    & \textbf{98.44} & 99.44      & \textbf{99.31} & \textbf{100.0} \\
\hline
\end{tabular}
\label{tab:3}
\end{minipage}\\
\begin{minipage}[t]{1\textwidth}
\caption{Results on finetuned generators.}
\vspace{-0.27cm}
\centering
\scriptsize
\setlength{\tabcolsep}{2.5pt}
\renewcommand{\arraystretch}{1.1}
\begin{tabular}{l|*{2}{c}|*{2}{l}|*{2}{l}|*{2}{l}}
\hline
\multirow{3}{*}{} &  \multicolumn{8}{c}{\textbf{Satellite image generation} }  \\
\cline{2-9}
& \multicolumn{2}{c|}{amusement\_park} & \multicolumn{2}{c|}{car\_dealership} &
 \multicolumn{2}{c|}{electric\_station} & \multicolumn{2}{c}{stadium}    \\\hline
  & Acc\% & AP\% & Acc\% & AP\% & Acc\% & AP\% & Acc\% & AP\%   \\
\hline
DIRE      & 50.49 & 85.96 & 50.19 & 84.49 & 50.40 & 83.85 & 50.58 & 94.11 \\
LGrad     & 48.85 & 35.98 & 50.00 & 42.38 & 50.15 & 47.30 & 50.02 & 43.46 \\
NPR       & 46.89 & 42.62 & 65.65 & 92.65 & 62.52 & 83.55 & 56.16 & 84.64 \\
FatFormer & 51.68 & 69.84 & 50.88 & 83.10 & 52.81 & 78.68 & 51.66 & 83.76 \\
RINE      & 69.07 & 82.59 & 51.97 & 50.15 & 64.19 & 73.99 & 68.16 & 78.23 \\
CoDE      & 86.66 & 82.00 & 88.40 & 83.67 & 84.50 & 79.53 & 78.37 & 73.16 \\
C2P-CLIP  & 58.00 & 75.82 & 76.60 & 96.21 & 71.71 & 87.63 & 58.74 & 93.90 \\
AIDE      & 52.90 & 57.74 & 61.71 & 84.69 & 61.11 & 75.62 & 52.32 & 72.37 \\
BFree     & \uwave{95.48} &	\uwave{100.0} &	\uwave{95.57} &	\uwave{99.97} &	\uwave{90.75} & \uwave{99.95}	& \uwave{88.95}	&\uwave{99.98}\\\hline
Ours\_VAE & 99.33         & \textbf{100.0} & 96.43         &99.96        & 96.08         & 99.87          & 96.50        & 99.97 \\
Ours\_VQ  & 96.77 & \textbf{100.0} & 87.52         &\textbf{99.98}&	90.99         &	\textbf{99.89} & 98.89        &	\textbf{99.99}\\
Ours\_Diffusion & \textbf{99.67}       & 99.97          & 97.94         & 99.62        &	96.50         &	99.17          & 97.34        & 99.43 \\
Ours\_Sparse &  99.61     & 99.99          & \textbf{98.91}  & 9.994        &	\textbf{98.55}&	99.82          &\textbf{99.59}& 99.98 \\\hline
\end{tabular}
\label{tab:4}
\end{minipage}
\end{table*}
\subsection{Results across different category of generators} \label{sec:5.1}

% \begin{figure*}[t]
% \begin{minipage}[b]{0.49\textwidth}
% \vbox{
%  \includegraphics[width=\linewidth]{bar_sd.jpeg}
%  \caption{Performance comparison of our model trained on the final component of SD2}\label{fig:4}
%  }
% \end{minipage}
% \hfill
% \begin{minipage}[b]{0.49\textwidth}
% \vbox{
%  \includegraphics[width=\linewidth]{bar_HiDream.jpeg}
%  \caption{Performance comparison of our model trained on the final component of HiDream}\label{fig:5}
%  }
% \end{minipage}
% \hfill
% \begin{minipage}[b]{0.24\textwidth}
% \includegraphics[width=.95\linewidth]{tsne.png}
%     \caption{Feature visualization of the three final components.}
%     \label{fig:6}
% \end{minipage}
% \end{figure*}

\textbf{Testing set}
(1) Synthbuster \cite{bammey2023synthbuster} is a challenging benchmark comprising 1,000 high-resolution real images from RAISE-1K \cite{dang2015raise} and 1,000 fake images generated by each of nine models: SD1.3, SD1.4, SD2, SDXL, DALL·E2, DALL·E3, Glide, Adobe Firefly and MidJourney v5, using the corresponding 1000 prompts. Based on our proposed taxonomy, these subsets are classified into the VAE.decoder and Diffusion.denoiser categories, as summarized in Table 1. As the architectures of Firefly and MidJourney are not  disclosed, we treat them as wild generators and evaluate them in Section~\ref{sec:5.3}. 

Since the release of Synthbuster, additional generators have emerged that are not included in the benchmark, including VQ.decoder-based generators and new VAE.decoder-based generators such as HiDream, Flux-dev, and SD3.5. To incorporate these, we generate fake images using two autoregressive VQ.decoder generators, Emu3 and LlamGen, as well as HiDream, all based on the same 1,000 Synthbuster prompts.

(2) Although the AI-generated images in these sub-testing sets come from various generators, the real image set is the same across all subsets. In other words, the model’s classification performance on the real images affects the results for all subsets when computing balanced binary classification accuracy. Moreover, 1,000 images for both the real and fake sets constitute a limited sample size and offer limited diversity.
To evaluate more diverse semantic content and larger sample sizes, for Flux-dev and SD3.5, we use 9,600 images from the PASCAL VOC training split~\cite{pascal-voc-2007} as the real image set. For the fake image set, we generate 10,000 images from each generator using short descriptions produced by an LLM.
Since the detector identifies generator traces rather than comparing real–AI image pairs, the semantic content differences between PASCAL VOC and the generated images do not affect the evaluation. Instead, the ability of the detector to correctly classify real images containing many small objects, as in PASCAL VOC, is a key aspect of assessing its robustness.

\begin{figure*}[t]
\vspace{-0.1cm}
\begin{minipage}[t]{1\textwidth}
    \includegraphics[width=1\textwidth]{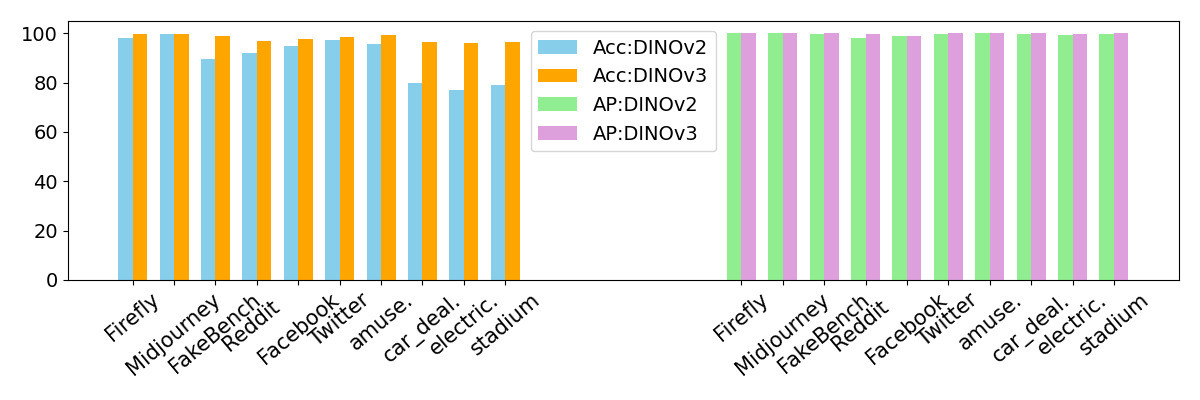}
    \vspace{-0.8cm}
    \caption{Performances of our detectors with different backbones}
    \label{fig:7}
\end{minipage}
\end{figure*}

\vspace{0.5em}
\noindent \textbf{Results} Table 1 %~\ref{tab:2} 
reports detection results across generator categories. The best results are highlighted in bold, and the best baseline results are indicated with a wavy underline. It is evident that robustness is a key challenge, especially as new generators continue to emerge. DIRE and LGrad, which were trained on GAN-based data, perform poorly on most modern generators. However, for Flux and SD3.5, LGrad achieves 81.91\% accuracy and 91.81\% AP. Similarly, RINE, benefiting from CLIP-based features, performs well on some generators, such as LlamaGen (90.65\% accuracy) and SD3.5 (96.49\%).
All baselines published before 2025 struggled to detect DALL·E3 images, a challenge that persisted until BFree was introduced. BFree remains the strongest baseline but still shows limited robustness on Glide, Flux, and HiDream generations. Notably, BFree produces high AP scores even when its prediction accuracy is low. For example, it achieves 69.62\% / 96.62\% (accuracy / AP) on Flux images, and 68.61\% / 93.81\% on Glide images. This discrepancy indicates that, although the model ranks positive samples effectively, its thresholded predictions may incorrect, which may pose challenges in applications that rely on well-calibrated prediction confidence.

In contrast, our approach performs strongly within each category and generalizes effectively across categories, consistently achieving higher accuracy on various subsets. Remarkably, the detector trained on the sparse set, using only 100 fake samples from each of the three components (300 fake and 300 real images in total), achieves performance comparable to our three models trained on individual components with the full MS-COCO dataset. This small training set provides further evidence of the generalization capability of our approach and offers a practical advantage, as generating large-scale training images is costly.

\subsection{Results on wild generations} \label{sec:5.3}

\textbf{Testing set}
(1) \textbf{Unknown architectures—Firefly and Midjourney} subsets from Synthbuster. Both are commercial generators lacking open-source code or public documentation of their architectures or components.
(2) \textbf{Mixed generators—FakeBench}\cite{li2025fakebench}. %~\footnote{\url{https://github.com/Yixuan423/FakeBench}}}
It contains 3,000 real images and 3,000 fake images generated by multiple models.
(3) \textbf{In-the-wild generation—WildRF}\cite{cavia2024real}. %~\footnote{\url{https://vision.huji.ac.il/ladeda/}}} 
It comprises three subsets: Reddit, Facebook, and Twitter, each containing real and synthetic images sourced from social media platforms with unknown origins.

\vspace{0.5em}
\noindent \textbf{Results} Table~\ref{tab:3} reports the results. Baseline methods exhibit inconsistent results across benchmarks. For instance, RINE performs well on Firefly (92.10\% accuracy) but drops sharply on Midjourney (55.45\%) and Facebook (51.56\%). BFree performs well on most subsets but degrades on FakeBench. This instability underscores the challenge of real-world generalization, particularly in true zero-shot scenarios where no prior information about the generators is available. In contrast, our detector consistently surpasses all baselines and maintains robust, stable performance across all subsets, demonstrating that leveraging traces from the final component yields strong generalization to diverse architectures and challenging real-world conditions.

\subsection{Results on domain-specific fine-tuned generators}

\textbf{Testing set}
SatelliteDiffusion\cite{khannadiffusionsat} dataset is generated using a satellite imagery model fine-tuned from SD2 and conditioned on satellite metadata. Real images are sourced from high-resolution satellite imagery, while synthetic counterparts are produced using textual descriptions of the real samples. Four subsets: \textit{amusement park}, \textit{car dealership}, \textit{electric substation}, and \textit{stadium} are tested, each including around 2,000 real and 2,000 generated images.

\vspace{0.5em}
\noindent \textbf{Results} Table~\ref{tab:4} presents the comparison. Our models achieve results closely matching those reported for the original SD2 evaluations in Table 1, demonstrating that the detection approach remains effective for generators fine-tuned on domain-specific images and conditions. 
Baseline models such as CoDE exhibit a similar phenomenon: on SD2, it achieves 88.05\% accuracy, while on fine-tuned subsets such as \textit{amusement\_park} and \textit{electric\_station}, the accuracy slightly decreases to 86.66\% and 84.50\%, respectively. On the \textit{stadium} subset, both CoDE and the best baseline, BFree, experience larger drops in performance.

\subsection{Ablation study}
\textbf{Different backbones} To evaluate the impact of the backbone on our detector, we train an additional model using the pretrained DINOv2 backbone. The comparison of performance between DINOv2 and DINOv3 is shown in Fig.~\ref{fig:7}. The results show that the newer backbone provides a significant improvement in accuracy. For AP, since the margin between the logits of positive and negative images is sufficiently large with DINOv2, using the newer backbone has little effect.

\vspace{0.5em}
\noindent\textbf{Final component vs. full pipeline}
To verify whether traces from the final component alone are sufficient to detect images generated by the full pipeline, we train an additional detector on images constructed from the final component of HiDream-I1 on the MS-COCO dataset using Algorithm~\ref{alg:1}. To evaluate its performance on full-pipeline outputs, we generate test images following Section~\ref{sec:5.1} using the 1,000 Synthbuster prompts. The detector achieves 90.5\% accuracy and 99\% AP, demonstrating that training on the generator’s final component generalizes effectively to its full outputs.

\section{Conclusion}

This study shows that leveraging the final components of modern image generators offers an effective strategy for improving deepfake detection generalization. By contaminating real images through these components and training on minimal representative samples, the proposed detector achieves superior performance across diverse unseen generators, highlighting the value of insights for robust detection.

\normalfont

%% file: X_suppl.tex
\clearpage
\setcounter{page}{1}
\setcounter{section}{0} 
\maketitlesupplementary

\section{Discussion on Property 2}
\label{sec:proof}

For convenience, the paired gradient $\hat{\nabla}_\theta^{(P)}$ can be rewritten as:
\small\[
\hat{\nabla}_\theta^{(P)} = -\frac{1}{|\mathcal{B}_t|} \sum_{i=1}^{|\mathcal{B}_t|} \Big[ \nabla_\theta \log \theta(\mathbf{x}_i) + \nabla_\theta \log (1 - \theta(\varphi(\mathcal{E}(\mathbf{x}_i)))) \Big].
\]
Define
\small\[
g_i = \nabla_\theta \log \theta(\mathbf{x}_i), \quad
g_i' = \nabla_\theta \log (1 - \theta(\varphi(\mathcal{E}(\mathbf{x}_i)))),
\]
and let
\small\[
G = \{ g_i \}_{i=1}^{|\mathcal{B}_t|}, \quad G' = \{ g_i' \}_{i=1}^{|\mathcal{B}_t|}.
\]
Under this notation, $\hat{\nabla}_\theta^{(P)}$ becomes
\small\[
\hat{\nabla}_\theta^{(P)} = -\frac{1}{|\mathcal{B}_t|} \sum_{i=1}^{|\mathcal{B}_t|} \big[ g_i + g_i' \big].
\]
Since the elements of $\mathcal{B}_t$ are sampled independently, each $\mathbf{x}_i$ is independent of all other samples in $\mathcal{B}_t$. Consequently, each gradient $g_i$ is independent of all other elements in $G$ except itself, and the same holds for $g_i'$ within $G'$. Furthermore, $g_i$ is independent of all elements of $G'$ except for its paired counterpart $g_i'$, and vice versa.

Using the fact that independent random variables have zero covariance, the variance of $\hat{\nabla}_\theta^{(P)}$ is:
\small\[
\begin{aligned}
\text{Var}(\hat{\nabla}_\theta^{(P)}) 
&= \frac{1}{|\mathcal{B}_t|} \sum_{i=1}^{|\mathcal{B}_t|} \Big[ \text{Var}(g_i) + \text{Var}(g_i') + 2 \text{Cov}(g_i, g_i') \Big] \\
&= \frac{1}{|\mathcal{B}_t|} \Big[ \text{Var}(G) + \text{Var}(G') + \text{Cov}(G, G') \Big],
\end{aligned}
\]
where
\small\[
\text{Var}(G) = \sum_{i=1}^{|\mathcal{B}_t|} \text{Var}(g_i), \quad
\text{Var}(G') = \sum_{i=1}^{|\mathcal{B}_t|} \text{Var}(g_i'), \]
\small\[
\text{Cov}(G,G') = 2 \sum_{i=1}^{|\mathcal{B}_t|} \text{Cov}(g_i, g_i').
\]
In contrast, the variance under independent sampling is
\small\[
\text{Var}(\hat{\nabla}_\theta^{(I)}) = \frac{1}{|\mathcal{B}_t|} \text{Var}(G) + \frac{1}{|\mathcal{B}_t|} \text{Var}(G').
\]
The only difference between the two variances is the covariance term $\text{Cov}(G, G')$, which aggregates the individual covariances between $g_i$ and $g_i'$.  
Our key observation is that $\mathbf{x}_i$ and its transformed version $\mathbf{x}_i' = \varphi(\mathcal{E}(\mathbf{x}_i))$ are highly similar, and thus their gradients $g_i$ and $g_i'$ tend to be positively correlated. Therefore, $\text{Cov}(g_i, g_i') > 0$, implying $\text{Cov}(G, G') > 0$. As a result,\small\[
\text{Var}(\hat{\nabla}_\theta^{(P)}) > \text{Var}(\hat{\nabla}_\theta^{(I)}).
\] This confirms our claim that the gradient estimator achieves lower variance when mini-batches are sampled independently from $\mathcal{D}$ and $\mathcal{D}'$.

We empirically compare pairwise training with independent training. Using the training set \textit{VAE}, we train detectors with identical model architectures, hyperparameters, and training data. The only difference lies in how mini-batches are constructed: pairwise training loads matched pairs ($\mathbf{x}$, $\mathbf{x}'$) into the same mini-batch, whereas independent training samples $\mathbf{x}$ and $\mathbf{x}'$ independently and loads them separately, while ensuring that each mini-batch contains the same number of positive samples $\mathbf{x}$ and negative samples $\mathbf{x}'$. Table A reports the results. We copy the baseline numbers and \textit{Ours\_vae} results from Table 1 of the main paper, along with the additional row for pairwise training. The average accuracy of pairwise training is 97.22\%, compared to 99.22\% for independent training—a 2\% gap. This empirical observation suggests that pairwise training does not provide an advantage over independent training, consistent with our theoretical analysis above.

\begin{figure*}[t]
\begin{minipage}[b]{1\textwidth}
\centering
\addtocounter{table}{1}
%\refstepcounter{table}
\qquad \textbf{Table A}. Comparison between pairwise training and independent training.  \\[4pt]
\scriptsize
\setlength{\tabcolsep}{2pt}
\renewcommand{\arraystretch}{1.1}
\begin{tabular}{l|*{2}{l}|*{2}{l}|*{2}{l}|*{2}{l}|*{2}{l}|*{2}{l}}
\hline
\multirow{3}{*}{} & \multicolumn{6}{c|}{\textbf{diffusion.denoiser based}} & \multicolumn{4}{c|}{\textbf{vq.de-tokenizer based}} & \multicolumn{2}{c}{\textbf{vae.decoder}} \\
\cline{2-13}
& \multicolumn{2}{c|}{DALL·E 2} & \multicolumn{2}{c|}{DALL·E 3} & \multicolumn{2}{c|}{Glide} & \multicolumn{2}{c|}{Emu3} & \multicolumn{2}{c|}{LlamaGen} & \multicolumn{2}{c}{HiDream} \\
\cline{2-13}
& Acc\% & AP\% & Acc\% & AP\% & Acc\% & AP\% & Acc\% & AP\% & Acc\% & AP\% & Acc\% & AP\% \\
\hline
DIRE      & 56.45 & 67.05 & 50.35 & 53.53 & 49.50 & 49.27 & 53.65 & 56.73 & 54.05 & 59.26 & 51.40 &	51.21 \\
LGrad     & 55.65 & 55.73 & 47.65 & 40.41 & 47.45 & 51.83 & 54.80 & 48.03 & 60.70 & 63.08 & 40.05 &	39.06\\
NPR       & 49.25 & 50.18 & 50.35 & 50.48 & 48.80 & 51.43 & 51.70 & 53.05 & 52.25 & 55.48 & 47.75 &	45.51\\
FatFormer & 54.12 & 56.02 & 35.88 & 33.95 & 67.77 & 75.75 & 63.72 & 66.10 & 77.21 & 87.73 & 34.48 &	31.28\\
RINE      & 85.90 & 94.05 & 42.35 & 30.83 & 63.95 & 72.09 & 62.10 & 69.68 & 90.65 & 98.36 & 44.15 &	39.20\\
CoDE      & 54.60 & 54.26 & 73.15 & 72.76 & 80.75 & 80.36 & 77.45 & 77.06 & 90.50 & 90.11 & 71.45 &	71.06\\
C2P-CLIP  & 55.60 & 76.87 & 63.20 & 90.45 & {86.65} & {96.53} & 68.70 & 90.60 & 83.80 & 97.85 & 51.20 & 51.02\\
AIDE      & 59.75 & 42.26 & 14.15 & 31.34 & 59.75 & 76.71 & 62.50 & 56.86 & 63.25 & 84.55 & 48.30 & 37.95\\
BFree     & {93.30} & {99.16} & {96.22} & {99.64} & 68.61 & 93.81 & {99.05} & {99.97} & {99.15} & {99.98} & {81.40} & {93.19} \\\hline
Ours\_VAE & 99.60 & {100.0} & 99.70   & 99.99          & {98.85} & {99.92} & 99.70      & 99.99   & 99.70 & 99.99  & {99.15}&{99.94}         \\
Ours\_VAE\_pair  & 98.10  & 99.98     & 98.20       & 99.70  &  98.05   & 99.91  &  98.20   & 99.98    & 98.20 & 99.98 &97.55 &  99.27   \\
\hline
\end{tabular}
\label{tab:2}
\end{minipage}
\begin{minipage}[b]{1\textwidth}
\centering 
\scriptsize
\setlength{\tabcolsep}{2pt}
\renewcommand{\arraystretch}{1.1}
\begin{tabular}{l|*{2}{l}|*{2}{l}|*{2}{l}|*{2}{l}||*{2}{l}|*{2}{l}}
\hline
\multirow{3}{*}{} & \multicolumn{12}{c}{\textbf{vae.decoder based}} \\
\cline{2-13}
& \multicolumn{2}{c|}{SD1.3} & \multicolumn{2}{c|}{SD1.4} & \multicolumn{2}{c|}{SD XL} & \multicolumn{2}{c||}{SD2} & \multicolumn{2}{c|}{Flux} & \multicolumn{2}{c}{SD 3.5} \\
\cline{2-13}
& Acc\% & AP\% & Acc\%& AP\% & Acc\% & AP\% & Acc\%& AP\%& Acc\%& AP\%& Acc\% & AP\% \\
\hline
DIRE      & 48.75 & 43.24 & 48.65 & 43.38 & 50.00 & 58.90  & 47.45 & 40.37 & 61.66 & 73.10 & 55.65 & 58.00\\
LGrad     & 54.45 & 46.46 & 55.85 & 48.14 & 58.65 & 53.26  & 48.75 & 42.80 & 81.91 & 91.81 & 77.29 & 87.08 \\
NPR       & 50.70 & 51.05 & 50.60 & 51.69 & 52.45 & 55.54  & 43.50 & 46.23 & 90.00 & 92.34 & 88.04 & 89.82 \\
FatFormer & 52.97 & 53.60 & 53.72 & 54.07 & 66.67 & 75.43  & 51.57 & 54.41 & 55.32 & 89.83 & 75.86 & 95.39 \\
RINE      & 87.60 & 95.96 & 87.15 & 95.95 & 74.60 & 85.56  & 70.30 & 80.74 & 51.70 & 86.61 & 96.49 & {99.93}\\
CoDE      & 97.75 & 97.36 & 97.30 & 96.91 & 75.20 & 74.81  & 88.05 & 87.66 & 71.09 & 72.17 & 76.35 & 75.26\\
C2P-CLIP  & 75.15 & 92.93 & 76.70 & 93.03 & 77.70 & 94.67  & 69.15 & 92.30 & 53.96 & 94.58 & 80.88 & 98.87 \\
AIDE      & 57.60 & 60.25 & 56.70 & 60.14 & 62.70 & 63.98  & 53.45 & 42.09 & {98.50} & {99.80} & 96.45 & 99.72 \\
BFree   & {99.25} & {99.99} & {99.15} & {99.99} & {99.25} & {99.98} & {99.80} & {99.93} & 69.62 & 93.62 & {98.74}  &  {99.94} \\\hline
Ours\_VAE & 99.70 & {100.0} & 99.70 & {100.0} & 99.70          & {100.0} & 99.70 & {100.0}& 99.17          & 99.98          & {99.96} & {100.0} \\
Ours\_VAE\_pair  & 98.20  &  99.99  & 98.20   & 99.99  & 98.20  & 99.92 & 98.20 & 99.93 & 98.88   & 95.42   &  99.07 & 97.75 \\
\hline
\end{tabular}
\end{minipage}
\end{figure*}

\section{Evidence sources for categorization}
\label{cat}
Table B summarizes the sources used to determine the final-component category of each generator. For all open-source models, we list the exact commit version, together with the file paths and line numbers that invoke the final component. DALL·E 2 provides only documentation rather than code, and DALL·E 3 has neither detailed documentation nor public source code. Among the listed generators, Flux-1-dev has two widely used official implementations: the Diffusers version operates in latent space and uses a VAE decoder as its final component, while the Black Forest Labs implementation performs generation directly in pixel space. A similar situation appears in DeepFloyd IF: its Stage III super-resolution step has two variants—one based on the Stable Diffusion ×4 latent upsampler, and another implemented in pixel space. Moreover, although MAR is grouped under the “denoiser” category, its final step is a decoder network structurally similar to a U-Net denoiser but without any noise addition or removal. 

\section{Baseline models}
\label{baseline}

As none of the baseline methods report results on our testing sets, we re-evaluate all of them using their publicly released pretrained models and default hyperparameters. All baselines compute Accuracy and AP using the same functions from \texttt{\hyphenchar\font=`\- sklearn.metrics}: \texttt{\hyphenchar\font=`\- accuracy\_score} and \texttt{\hyphenchar\font=`\- average\_precision\_score}. The only exception is CoDE, which does not include AP computation in its evaluation code; for this case, we add the same AP calculation used by the other baselines. Below, we provide the links to the testing code and pretrained weights. For each baseline, we use the default hyperparameters specified in the authors’ example scripts within their repositories. 

\begin{itemize}
\item \textbf{DIRE} (ICCV 2023): Testing code from \url{https://github.com/ZhendongWang6/DIRE}. We use the pretrained model \texttt{\hyphenchar\font=`\- lsun\_adm.pth} downloaded from the provided checkpoints. 
\item \textbf{LGrad} (CVPR 2023): Testing code from \url{https://github.com/chuangchuangtan/LGrad}, with pretrained weights \texttt{\hyphenchar\font=`\-  LGrad-4class-Trainon-Progan\_car\_cat\_chair\_horse.pth}.
\item \textbf{NPR} (CVPR 2024): Testing code from \url{https://github.com/chuangchuangtan/NPR-DeepfakeDetection}, with pretrained weights \texttt{\hyphenchar\font=`\- model\_epoch\_last\_3090.pth}.
\item \textbf{FatFormer} (CVPR 2024): Testing code from \url{https://github.com/Michel-liu/FatFormer}, using \texttt{\hyphenchar\font=`\- fatformer\_4class\_ckpt.pth}.
\item \textbf{RINE} (ECCV 2024): Testing code from \url{https://github.com/mever-team/rine}, using \texttt{\hyphenchar\font=`\- model\_ldm\_trainable.pth}.
\item \textbf{CoDE} (ECCV 2024): Testing code from \url{https://github.com/aimagelab/CoDE}, with the pretrained classifier available at 
\url{https://huggingface.co/aimagelab/CoDE/tree/main/sklearn/linear\_tot\_classifier\_epoch-32.sav}.
\item \textbf{C2P-CLIP} (AAAI 2025): Testing code from \url{https://github.com/chuangchuangtan/C2P-CLIP-DeepfakeDetection}, using the released weights at \url{https://www.now61.com/f/95OefW/C2P\_CLIP\_release\_20240901.zip/C2P\_CLIP\_release\_20240901.pth}.
\item \textbf{AIDE} (ICLR 2025): Testing code from \url{https://github.com/shilinyan99/AIDE}, using \texttt{\hyphenchar\font=`\- progan\_train.pth}.
\item \textbf{BFree} (CVPR2025): Testing code from \url{https://github.com/grip-unina/B-Free}, using weights \url{https://www.grip.unina.it/download/prog/B-Free/weights/BFREE\_dino2reg4.zip}.
\end{itemize}

\begin{table*}[!htbp]
\sloppy
\centering
\refstepcounter{table}
\qquad Table B. Sources for generator categorization. \\[4pt]
\scriptsize
\scriptsize
\setlength{\tabcolsep}{3pt}
\renewcommand{\arraystretch}{1.3}
\begin{tabular}{p{1.2cm}|p{1.2cm}|p{4.3cm}|>{\raggedright\arraybackslash}p{4.2cm}|p{6cm}}
\hline
Name & Category & Source & Evidence \\\hline  
Stable Diffusion 3.5    & vae.decoder  & \url{https://github.com/Stability-AI/sd3.5}                                           & Commit 106db06 \newline sd3\_infer.py : Line 495 
& \verb|image = self.vae_decode(sampled_latent)|  \\\hline
Flux 1-dev              & vae.decoder  & \url{https://github.com/huggingface/diffusers/tree/main/src/diffusers/pipelines/flux} & Commit 5e181ed \newline pipeline\_flux.py : Line 1006 
& \texttt{\hyphenchar\font=`\- image = self.vae.decode(latents, return\_dict=False)[0]} \\\cline{2-5}
& diffusion. denoiser & \url{https://github.com/black-forest-labs/flux} & Commit 57ce405 \newline src/flux/sampling.py : Line 351 & \texttt{\hyphenchar\font=`\- img = img + (t\_prev - t\_curr) * pred}\\\hline
HiDream-I1              & vae.decoder  & \url{https://github.com/HiDream-ai/HiDream-I1}                            & Commit 3519729 \newline hi\_diffusers/pipelines/hidream\_image/ pipeline\_hidream\_image.py : Line 724 
& \texttt{\hyphenchar\font=`\- image = self.vae.decode(latents, return\_dict=False)[0]} \\\hline
Stable diffusion xl     & vae.decoder   & \url{https://github.com/huggingface/diffusers/tree/main/src/diffusers/pipelines/stable_diffusion_xl &}  & Commit 0d1c5b0 \newline pipeline\_stable \_diffusion\_xl.py : Line 1292 
& \texttt{\hyphenchar\font=`\- image = self.vae.decode(latents, return\_dict=False)[0]} \\\hline
CogView4            & vae.decoder      & \url{https://github.com/huggingface/diffusers/blob/main/src/diffusers/pipelines/cogview4}  &  Commit 0d1c5b0 \newline pipeline\_cogview4.py : Line 673
& \texttt{\hyphenchar\font=`\- image = self.vae.decode(latents, return\_dict=False, generator=generator)[0]}\\\hline
DeepFloyd IF        & vae.decoder      & \url{https://github.com/deep-floyd/IF/} & Commit af64403 \newline deepfloyd\_if/modules/stage\_III\_sd\_x4.py : Line 80  
& \texttt{\hyphenchar\font=`\- available\_models = ['stable-diffusion-x4-upscaler'] ... \newline images = self.model(**metadata).images } \\
& & & \\
&  & \url{diffusers/src/diffusers/pipelines/stable_diffusion} & Commit a4df8db \newline pipeline\_stable \_diffusion\_upscale.py : Line 798 & \texttt{\hyphenchar\font=`\- image = self.vae.decode(latents / self.vae.config.scaling\_factor, return\_dict=False)[0]} \\\cline{2-5}
& diffusion. denoiser & \url{https://github.com/deep-floyd/IF/} & Commit af64403 \newline deepfloyd\_if/modules/stage\_III.py:Line 26 & \texttt{\hyphenchar\font=`\- return super().embeddings\_to\_image(...)} \\
& & & Commit af64403 \newline deepfloyd\_if/modules/base.py : Line 194 & \texttt{\hyphenchar\font=`\- sample = diffusion.p\_sample\_loop(...)} \\\hline
Kandinsky 3.0      & vq.de-tokenizer       & \url{https://github.com/ai-forever/Kandinsky-3}  & Commit 10db67a \newline kandinsky3/movq.py: Line 420,421 & \texttt{\hyphenchar\font=`\- decoder\_input = self.post\_quant\_conv(quant)   decoded = self.decoder(decoder\_input, quant)} \\\hline
Emu3               & vq.de-tokenizer       & \url{https://github.com/baaivision/Emu3} & Commit 1a43ee6 \newline emu3/mllm/processing\_emu3.py : Line 198, 199& \texttt{\hyphenchar\font=`\- doc=self.tokenizer.decode(*args,**kwargs) return self.multimodal\_decode(doc)}\\\hline
JanusPro           & vq.de-tokenizer       & \url{https://github.com/deepseek-ai/Janus} &  Commit 659982f\newline janus/models/ vq\_model.py: Line 506,507 & \texttt{\hyphenchar\font=`\- quant\_b = self.quantize.get\_codebook\_entry() dec = self.decode(quant\_b)} \\\hline
LlamaGen           & vq.de-tokenizer       & \url{https://github.com/FoundationVision/LlamaGen} & Commit 5a50452\newline autoregressive/sample/
sample\_t2i.py : Line 121 & \texttt{\hyphenchar\font=`\- samples = vq\_model.decode\_code(index\_sample, qzshape)} \\\hline
Infinity           & vq.de-tokenizer       & \url{https://github.com/FoundationVision/Infinity} & Commit 3ab8e0c \newline infinity/models/infinity.py : Line 636  & \texttt{\hyphenchar\font=`\- img = vae.decode(summed\_codes.squeeze(-3))}\\\hline
VAR                & vq.de-tokenizer       & \url{https://github.com/FoundationVision/VAR}  & Commit a5cf0a1 \newline models/var.py : Line 190 & \texttt{\hyphenchar\font=`\- return self.vae\_proxy[0].fhat\_to\_img(f\_hat) // self.vae\_proxy: Tuple[VQVAE]} \\\hline
MAR                & decoder       & \url{https://github.com/LTH14/mar}  & Commit fe1de72 \newline models/mar.py : Line: 225 & \texttt{\hyphenchar\font=`\- x = block(x)} \\\hline
DALL·E 2           & diffusion. denoiser       & Document: Hierarchical Text-ConditionalImage Generation with CLIP Latents  &  Page 4 & ``we train two diffusion upsampler models: one to upsample images from 64×64 to 256×256 resolution, and another to further upsample those to 1024×1024 resolution" \\\hline
GLIDE              & diffusion. denoiser       & \url{https://github.com/openai/glide-text2im} & Commit 1f791b8 \newline glide\_text2im/text2im\_model.py: Line 167 & \texttt{\hyphenchar\font=`\- return super().forward(x, timesteps, **kwargs)} \\\hline
PixelFlow          & diffusion. denoiser       & \url{https://github.com/ShoufaChen/PixelFlow} & Commit eeabc08 \newline pixelflow/pipeline\_pixelflow.py : Line 253 & \texttt{\hyphenchar\font=`\- latents = self.scheduler.step( model\_output=noise\_pred,sample=latents)} \\\hline
Reflection- Flow     & diffusion. denoiser       & \url{https://github.com/Diffusion-CoT/ReflectionFlow} & train\_flux
/sample.py : inference-time refine by black-forest-labs/FLUX.1-dev & Flux-1-dev (pixel-space implmentation; see note above) \\\hline
DDPM               & diffusion. denoiser       & \url{https://github.com/hojonathanho/diffusion} & Commit c461299\newline diffusion\_tf/diffusion\_utils\_2.py : Line 211 & \texttt{\hyphenchar\font=`\- self.p\_sample(..., x=img\_, ...)}\\\hline
DDIM               & diffusion. denoiser       & \url{https://github.com/huggingface/diffusers/tree/main/src/diffusers/pipelines/ddim} & Commit a4df8db\newline pipeline\_ddim.py : Line 152  & \texttt{\hyphenchar\font=`\- image = self.scheduler.step(...)} \\\hline
GigaGAN            & one-stop. generator       & \url{https://github.com/lucidrains/gigagan-pytorch} & Commit 0806433 \newline gigagan\_pytorch/gigagan\_pytorch.py: Line 2167, 2169 &  \texttt{\hyphenchar\font=`\- model = self.G\_ema if self.has\_ema\_generator else self.G \newline return model(*args, **kwargs)} \\\hline
JetFormer          & one-stop. generator       & \url{https://github.com/google-research/big_vision} & Commit 9c006bf\newline big\_vision/models/proj/jetformer/ jetformer.py: Line 481 & \texttt{\hyphenchar\font=`\- prelogits, decoder\_out = self.decoder(x, ...)}\\
\hline
\end{tabular}
\label{tab:6}
\end{table*}

\section{Datasets construction}
\label{dataset}

Based on \textbf{Property 2}, we generate the constructed image sets ${\mathbf{x}'}$ offline and save them before training begins. For each of the three representative final components---\textit{Stable Diffusion 2.1,JanusPro and PixelFlow}, the corresponding constructed training set is produced using its well-trained, optimized encoder, following the guidance of \textbf{Property 1}. Below, we list the real images and the pretrained $\mathcal{E}^*$ and $\varphi^*$ used in our experiments:
\begin{itemize}
\item \textbf{VAE training set} (Sec 5.1): The real images are from the full MS-COCO 2014 dataset (118k images). The pretrained VAE used to generate negative samples is the VAE from \url{https://huggingface.co/stabilityai/stable-diffusion-2-1}.

\item \textbf{VQ training set} (Sec 5.1): The pretrained model used for converting real images is JanusPro-7B \url{https://huggingface.co/deepseek-ai/Janus-Pro-7B}. Due to its fixed patch size, Janus only supports a resolution of 378×378. Therefore, we filter out images smaller than this size and center-crop the remaining MS-COCO images to 378×378. 
\item \textbf{Diffusion training set} (Sec 5.1): We use PixelFlow \url{https://huggingface.co/ShoufaChen/PixelFlow-Text2Image/tree/main/model.pt}, whose last stage is pretrained to upscale images from 128×128 to 512×512. For real images in MS-COCO, we crop them to 512×512 whenever the resolution allows. We do not downsample and re-upsample the real images; instead, we directly use real images that match the output resolution. 
\item \textbf{HiDream training set} (Sec 5.5): The pretrained VAE weights are from HiDream-I1-Full \url{https://huggingface.co/HiDream-ai/HiDream-I1-Full}.
\end{itemize}

For the testing set, we use the full sets from the benchmarks Synthbuster, FakeBench, and WildRF. In Synthbuster, there are 1,000 high-resolution real images from RAISE-1K, along with 1,000 fake images generated by each of nine models: DALL·E2, DALL·E3, Adobe Firefly, MidJourney v5, SD1.3, SD1.4, SD2, SDXL, and Glide. The corresponding columns in Tables 1 and 2 correspond to these subsets. To include more recent generators, we additionally construct the following test sets used in the experiments reported in the main paper:
\begin{itemize}
\item \textbf{HiDream, Emu3, and LlamaGen} (Table 1): Real images are the same as in Synthbuster. Fake images are generated from the same 1,000 prompts used for the original nine generators (listed in \texttt{\hyphenchar\font=`\- prompts.csv} in \url{https://zenodo.org/records/10066460/files/synthbuster.zip}). Generation parameters and image resolutions follow the defaults in each repository, as listed in Table B.
\item \textbf{Flux-1-dev and SD3.5} (Table 1): We generate additional large-scale test sets from SD3.5 (\url{https://github.com/Stability-AI/sd3.5}) and Flux-1-dev (\url{https://github.com/huggingface/diffusers/tree/main/src/diffusers/pipelines/flux}), with 10,000 fake images each.
\item \textbf{SatelliteDiffusion testing set} (Table 3): The real images comprise all images from four categories of the fMoW validation set (\url{https://github.com/fMoW/dataset}): amusement park (2,556 images) , car dealership (2,156 images), electric substation (2,626 images ), and stadium (2,575 images). Each real image has associated metadata including location (country). Fake images are generated from \url{https://github.com/samar-khanna/DiffusionSat} using prompts of the form: ``a fMoW satellite image of \{class\_name\} in \{country\}''. The number of generated images matches the number of real images for each class.
\end{itemize}